%% file: paper.tex
\documentclass[5p,twocolumn,times]{elsarticle}

\usepackage{amssymb,amsmath}

\usepackage{booktabs}
\usepackage{multirow} 
\makeatletter
\newcommand*\my@starttable[1][]{%
	  \@float{table}[#1]\footnotesize
		}
		\patchcmd{\table}{\@float{table}}{\my@starttable}{\PackageInfo{mysty}{Table environment patched successfully.}}{\PackageWarning{mysty}{Could not patch table environment.}}
\makeatother

\usepackage{listings}
\usepackage{xcolor}
\definecolor{codegray}{rgb}{0.5,0.5,0.5}

\lstdefinestyle{jsonstyle}{
	numberstyle=\tiny\color{codegray},
	stringstyle=\color{codepurple},
	basicstyle=\ttfamily\footnotesize,
	breakatwhitespace=false,         
	breaklines=true,                 
	captionpos=b,                    
	keepspaces=true,                 
	numbers=left,                    
	numbersep=5pt,                  
	showspaces=false,                
	showstringspaces=false,
	showtabs=false,                  
	tabsize=2
}

\usepackage[hidelinks]{hyperref}

\usepackage{caption}
\usepackage{subcaption}

\usepackage{tikz}
\usetikzlibrary{shapes.geometric, arrows, positioning, shadows}

\usepackage{flushend}

\journal{~}

\usepackage{xspace} 
\input{datastats}

\newcommand{\code}{\texttt}

\begin{document}
\begin{frontmatter}

	\title{Detecting Network-based Internet Censorship 
	via Latent Feature Representation Learning}
	\author[bc]{Shawn P. Duncan}

	\author[bc,gc]{Hui Chen\corref{cor1}}
	\affiliation[bc]{organization={Department of Computer \& Information Science, CUNY Brooklyn College},
            addressline={2900 Bedford Avenue}, 
            city={Brooklyn},
            postcode={11210}, 
            state={NY},
            country={USA}}
	\affiliation[gc]{organization={Department of Computer Science, CUNY Graduate Center},
            addressline={365 5th Avenue}, 
            city={New York},
            postcode={10016}, 
            state={NY},
            country={USA}}

	\begin{abstract}
		\input{abstract}
	\end{abstract}

	\begin{keyword}
		Internet censorship \sep 
		Internet censorship detection \sep
		Network-based Internet censorship detection \sep 
		Deep neural network \sep 
		Feature representation learning


	\end{keyword}

\end{frontmatter}

\input{intro}

\input{related}
\input{system}

\input{eval}

\input{threats}

\input{summ}

\input{ack}

\biboptions{sort&compress}
\bibliographystyle{elsarticle-num} 
\bibliography{paper}


\end{document}

%% file: datastats.tex
\newcommand\FLnrecords{1,369,071\xspace} 
\newcommand\FLntrain{1,197,937\xspace} 
\newcommand\FLnvalid{171,134\xspace}   
\newcommand\FLsamplingratio{10\%\xspace}
\newcommand\FLbatchsize{16\xspace}
\newcommand\FLlrbegin{0.003\xspace}
\newcommand\FLlrend{0.0001\xspace}

\newcommand\FLseqlen{7,069\xspace}

\newcommand\FLembsize{96\xspace}
\newcommand\FLembsizemin{96\xspace}
\newcommand\FLembsizemax{128\xspace}
\newcommand\FLenchidden{128\xspace}
\newcommand\FLencoutput{96\xspace}
\newcommand\FLdechidden{128\xspace}
\newcommand\FLdecinput{96\xspace}
\newcommand\FLdecoutput{7,069\xspace}
\newcommand\FLlossthreshold{20\%\xspace}

\newcommand\CDnrecords{1,711,339\xspace}
\newcommand\CDncensored{215,016\xspace}
\newcommand\CDnuncensored{653,481\xspace}
\newcommand\CDnunknown{842,842\xspace}
\newcommand\CDtrainratio{70\%\xspace}
\newcommand\CDvalidratio{10\%\xspace}
\newcommand\CDtestratio{20\%\xspace}

\newcommand\CDinputsize{96\xspace}
\newcommand\CDfirstsize{512\xspace}
\newcommand\CDsecondsize{256\xspace}
\newcommand\CDoutputsize{2\xspace}

%% file: abstract.tex
Internet censorship is a phenomenon of societal importance and attracts
investigation from multiple disciplines.  Several research groups, such as Censored Planet, have deployed large scale Internet
measurement platforms to collect network
reachability data.  However, existing studies generally rely on manually
designed rules (i.e., using censorship fingerprints) to detect 
network-based Internet censorship
from the data.  While this rule-based approach yields a high true positive
detection rate, it suffers from several challenges: it requires human
expertise, is laborious, and cannot detect any censorship not captured by the
rules.  Seeking to overcome these challenges, we design and evaluate a classification model based on  latent
feature representation learning and an image-based classification model to detect
network-based Internet censorship. To infer latent feature representations from
network reachability data, we propose a sequence-to-sequence autoencoder to
capture the structure and the order of data elements in the data. To estimate
the probability of censorship events from the inferred latent features, we rely on a densely connected
multi-layer neural network model.  Our image-based classification model encodes a network reachability data
record as a gray-scale image and classifies the image as censored or not using a
dense convolutional neural network.  We compare and evaluate both approaches using data
sets from Censored Planet via a hold-out evaluation.  Both
classification models are capable of detecting network-based Internet
censorship as we were able to identify instances of
censorship not detected by the known fingerprints. Latent feature representations likely encode more nuances in the data since the latent feature learning approach discovers  a greater quantity, and a more
diverse set, of new censorship instances.


%% file: intro.tex
\section{Introduction}
\label{sec:intro}

\begin{figure}[!htbp]
	\centering
	\includegraphics[width=0.9\columnwidth]{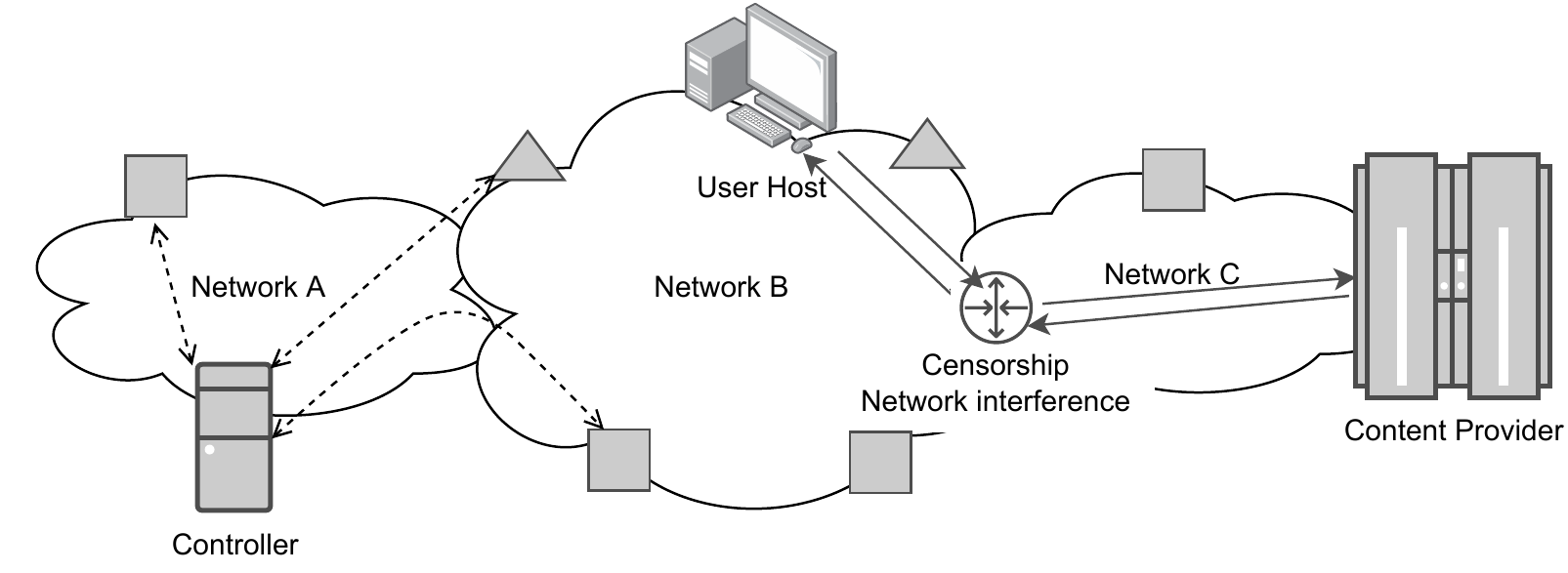}
	\caption{Network-based Internet censorship and measurement. Network-based
	Internet censorship takes place at neither the user host (or the application
	on the host) nor the content provider. To gauge such censorship, a measurement
	platform sets up vantage points, depicted as squares and triangles in the
	figure. The controller of the platform instructs the vantage point to
	send a network request to the content server or sends a request appearing
	from the vantage point to the content server. Although the existing platforms
	adopt more complex methods, a straightforward method to detect censorship, e.g., whether Network
	B censors a network request is to compare the network traces from multiple
	vantage points. Following that, we can design a fingerprint for the network
	trace for the censored vantage point and the content provider.} 
	\label{fig:nicensor}
\end{figure}

This study focuses on a particular type of Internet censorship called {\em
network-based Internet censorship}, i.e, the impairing or blocking of access to
online content and service {\em intentionally} by the network, a {\em third
party} between the user host and the online content and service
provider~\cite{aceto_internet_2015}, where an example of the online content can
be a blog post on the Web~\cite{zhu2013velocity} and that of the service can be
the Virtual Private Network (VPN) service~\cite{xue280012}.
Figure~\ref{fig:nicensor} illustrates this concept. This type of censorship
needs the control of neither the user hosts and applications nor the content and
service providers.  Rather it requires the access to the network components, such
as, routers and DNS servers between the user hosts and the content and service
providers. It is prevalent among those, such as a state actor, who seek to
block or impair access to politically, culturally, or religiously sensitive content
hosted on a server within or out of its jurisdiction without a need to tamper
with  the user hosts, the user applications, and the content servers.  

The network-based Internet censorship can lead to network
outages~\cite{dainotti2011analysis}.  Differentiating the censorship events
from other types of outages can thus benefit the design and implementation of
Internet network architecture, protocols, and
systems~\cite{aceto2018comprehensive, aceto_internet_2015}.  Understanding the
censorship mechanism can also help develop circumvention
techniques~\cite{winter2012great, chai2019importance, satija2021blindtls}.

Aside from these technical perspectives, there have been pursuits to gauge its
social, political, cultural, and religious
implications~\cite{shishkina2018internet, meserve2018google,
chen2000pornography}. These works reveal the complexity and the intricacy of Internet censorship, which sometimes challenges conventional wisdom
given its prevalence among countries, polities, cultures, and religions 
of a great spectrum~\cite{aryan2013internet, nabi2013anatomy, singh2017characterizing,
yadav2018light, ng2018detecting,
meserve2018google, bock2021even, padmanabhan2021multi, ververis2021understanding}.

In order to study censorship prevalence and mechanisms systematically, to understand this global phenomenon,
to improve the Internet infrastructure given
the prevalence of it, 
prior efforts have led to
the design, implementation, and deployment of large scale Internet measurement
platforms.  
Internet censorship
manifests as a network reachability problem, i.e., the users are unable to
access the content or the service, or are accessing the content or the service
at a severely degraded quality of service. These platforms are designed to collect
network measurement data about network reachability, 
and further, to detect Internet censorship events from the
reachability data~\cite{sundara_raman_censored_2020, niaki2020iclab,
filasto2012ooni}.
As shown in Figure~\ref{fig:nicensor}, these measurement platforms typically
set up vantage points, strategically or sometimes opportunistically selected
monitoring hosts on the Internet. The platforms typically have controller hosts
(that the platforms may name differently)
and the controllers have network requests sent to the content providers from or
appearing to originate from the vantage points. 

These measurement platforms produce a large amount of network
reachability data~\cite{sundara_raman_censored_2020, niaki2020iclab}.  There
are challenges to detecting censorship events from the reachability data, such as,
the platforms must process the amount of data, and they also need to find a way
to infer whether the reachability problem is intentional to differentiate a
censorship event from an ordinary network outage.  These works generally rely on a
rule-based approach to detect censorship, such as, by matching the measurement
data to censorship fingerprints that are usually purposefully designed regular
expressions~\cite{sundara_raman_censored_2020, niaki2020iclab}. The rule-based
approach has several advantages, such as, it is computationally efficient to
execute a rule, and a carefully designed rule is also unlikely to have a false
detection. However, the rule-based approach has several disadvantages. First,
it requires significant technical expertise.  Second, it is laborious to add
new rules.  Third, it cannot detect any Internet censorship not described by
existing rules. 

Machine learning-based approaches have found applications in software and
system quality assurance and security~\cite{qiao2020deep, lin2020software}.
An improvement over the rule-based approaches where the rules are pre-defined by human
experts manually, the learning-based approaches automatically learn ``rules''
or ``patterns'' in the relevant data. As a result, they can not only offer
better predictive performance than the rule-based systems, but also reduce
human efforts required to design rules for the rule-base systems.
Learning-based approaches are making inroads in both the design of censorship
mechanisms and the data analysis and detection of
censorship~\cite{gao2021machine, li2015predicting}.  Thus far, for
network-based Internet censorship detection, most prior work uses machine
learning algorithms in a semi-automated fashion. For instance, clustering
algorithms are used to group {\em block pages}, notices that content providers
respond to the users to explain the unavailability of requested
content~\cite{jones2014automated}.  A necessary condition to the learning-based
approaches (including clustering) is to express the network reachability data in
a form, i.e., a representation of data, suitable for machine learning
algorithms. This requires us to decide how we extract and represent features
from the data. One approach is to borrow features from natural language
processing, such as, to use features like page length and
term-frequency~\cite{jones2014automated, niaki2020iclab}.  Another is to borrow
approaches from image recognition, such as, to extract feature representations
as screenshots of block pages using ResNet50, a pre-trained convolutional deep
neural network~\cite{raman_measuring_2020}.

Limitations of the prior learning-based approaches for network-based Internet
censorship are two. First, the simple features like page length and
term-frequency do not sufficiently capture nuances in the network reachability
data, e.g., block pages, which results in a high false positive
rates~\cite{yadav2018light}. Second, the prior work primarily uses the learning
algorithms to aid the censorship detection rule design. One example is
FilterMap~\cite{raman_measuring_2020}.  It extracts feature representations from
screenshots of block pages using ResNet50 and clusters block pages into
approximately 200 groups of block pages.  Following this, the researchers
manually design detection rules (signatures of block pages) by examining each
group of block pages. Another example is ICLab~\cite{niaki2020iclab}.  It
computes the term-frequencies of HTML tags and the Locality Sensitive Hashing
index of the text in block pages, produces 48 block page clusters that are
similar either in HTML structures or in page text.  Again, the researchers
manually design new signatures for these clusters in the form of regular expressions. 

Despite these limitations, the prior work has led to the creation of hundreds
of network-based censorship fingerprints in the form of regular expressions and
discovered new censorship events~\cite{niaki2020iclab,
sundara_raman_censored_2020}. In addition, the work also results in the
accumulation of a large amount of reachability data labeled with censorship
types according to the signatures. Taking advantage of these and motivated
by the recent development in deep learning, we present and evaluate a 
censorship detection approach that is built upon well-known deep learning
techniques. The contributions of this work are as follows: 
\begin{enumerate}
	\item The work results in an ``end-to-end'' censorship detection system
		that automates the entire process of censorship detection.  On one
		end, the system automatically learns latent features from network
		reachability data records, which yields feature
		representations of the records. On the other end, from the feature
		presentations, we determine whether a network reachability problem is a
		censorship event via a classification model. The system therefore
		eliminates the need manually to design machine learning features and
		reduces the manual effort to detect new censorship events.  

	\item The network reachability data contains structures and
		pieces of some data elements appear in a particular order. Observing the
		characteristics of the data, we design the feature presentation learning 
		model to be a sequence-to-sequence model adapted from natural language processing. 

	\item The proposed classification model is a supervised learning model. Supervised
		learning models require a large set of labeled data to train, e.g., a
		known set of network measurement records that are labeled as censored or uncensored.  Our
		results show that the proposed supervised learning models trained using
		the block pages labeled by known censorship fingerprints 
		can recognize both known and new censorship
		events.  As a result, we hypothesize that there is additional information
		in the labeled block pages beyond what the signatures capture. 

	\item To evaluate the above end-to-end censorship detection approach and to 
		explore the design space, we design an alternative detection approach 
		that treats network reachability data records as gray-scale images
		and train a state-of-the-art deep learning image classification 
		model to recognize the images as
		censored or not. To compare the two proposed approaches, 
		we process a large collection of Censored Planet block page records.
		The end-to-end approach appears to have an advantage over the
		alternative due in part to the sequence-to-sequence feature representation
		learning.

	\item This work also results in a library of Python scripts that can be used
		for replication and future research. The authors make this library
		publicly available. 
\end{enumerate}

We structure the rest of this article as follows. To situate the proposed work,
we discuss related work in Section~\ref{sec:related}. We introduce our
proposal, the latent feature representation based approach and the image-based
alternative in Section~\ref{sec:design}. In Section~\ref{sec:eval}, we evaluate 
the proposed approaches. Our work isn't without limitations.
We dedicate Section~\ref{sec:threats} to a discussion of the limitations and
the threats to validity. Finally, we conclude this work in
Section~\ref{sec:summary}.

%% file: related.tex
\section{Related Work}
\label{sec:related}

\paragraph{Internet Censorship} 
Internet Censorship has been a global phenomenon, which has attracted a
sustained research interest and resulted in a plethora of literature.  
A significant body of the literature is to understand the prevalence of
censorship, such as, to measure the online content, the network protocols and
applications, and the geographical regions that are subject to
censorship~\cite{aryan2013internet, nabi2013anatomy, singh2017characterizing,
yadav2018light, ng2018detecting,
meserve2018google, bock2021even, padmanabhan2021multi, ververis2021understanding}.  
Another is to unpack the censorship mechanisms, such as,
IP-based blocking, TCP packet injection, Transparent proxy,  DNS manipulation,
and SNI filtering~\cite{aceto_internet_2015, chai2019importance, bock2021even,
niaki2020iclab}.  

This work is not aimed to expand these bodies of knowledge.  Instead, our focus
is about the design of the ``end-to-end'' network-based Internet censorship
detection system by leveraging the recent advances in deep learning and a wealth
of network reachability data collected by the large scale Internet measurement
platforms.

\paragraph{Censorship Measurement Platforms}
Large scale remote network measurement
platforms for studying Internet censorship have emerged over the last decade.~\cite{niaki2020iclab,
filasto2012ooni, sundara_raman_censored_2020, pearce2017augur,
pearce2017global, scott2016satellite, vandersloot2018quack, raman_measuring_2020,
jin2021understanding}. These platforms fall into two broad categories with regard to the measurement
methods employed and the selection and use of vantage points.

Some platforms
like OONI~\cite{filasto2012ooni}, ICLab~\cite{niaki2020iclab}, and
Disguiser~\cite{jin2021understanding} use a mixture of volunteer hosts, 
selected VPN servers, SOCKS Proxies, or RIPE Atlas servers as vantage 
points. From these vantage points the platforms make attempts to access a brand range
of content and services on the Internet, typically {\em like an
ordinary user would do}, and collect reachability data.  

Platforms like Censored
Planet~\cite{sundara_raman_censored_2020} take a different approach. Using systems like
Augur~\cite{pearce2017augur}, Iris~\cite{pearce2017global},
Satellite~\cite{scott2016satellite}, Quack~\cite{vandersloot2018quack}, and
HyperQuack~\cite{raman_measuring_2020}, these platforms rely on {\em network side-channels} to
gauge potential network censorship events. The distinction here is that
ordinary users would not access online content and services via these
side-channels; however, the platforms can select servers that support the
side-channel measurement mechanism globally as the vantage points.  These two
categories of platforms are complementary. For instance, Censored Planet can
select far more vantage points to provide a more complete geographical and
network coverage and continuity than OONI and
ICLab~\cite{sundara_raman_censored_2020}; however, the censorship mechanisms
that are applied to ordinary network traffic may not be always applicable to
the network side-channels, and as such, Censored Planet may not be able to
detect some Internet censorship events, which can lead to higher false negative
rate than OONI and ICLab~\cite{niaki2020iclab}.  

The platforms are the foundation and the motivation of this work. First, these
platforms are continuously collecting network measurement data, which provides
a foundation for the proposed data-driven Internet
censorship detection systems using deep neural networks. Second, the platforms have made available 
a large collection of censorship fingerprints, e.g., Censored Planet makes
available a tool to scan the collected data records to match a set of known
censorship fingerprints. Our system adopts a supervised learning model to
classify network reachability data records, and for this the fingerprints and
the tool are particularly useful to create labeled training data set 
for the classification model.
Third, some platforms use machine learning, particularly, unsupervised
learning to cluster network reachability data records to design new censorship
fingerprints~\cite{raman_measuring_2020, niaki2020iclab}, which is a 
laborious process and requires human expertise. This
observation motivates us to propose the end-to-end network-based
Internet censorship detection systems. 

For this work, we elect to use the Censored Planet data set to study the
proposed approaches to detect network-based Internet censorship.  Censored Planet
is distinct in that the
platform collects network measurement data on network side-channels.  The
censorship measured is therefore network-based censorship other than
host and server-based censorship~\cite{aceto_internet_2015} --- an exception may 
be censorship mechanism deployed on host-based network filters or firewalls.  

\paragraph{Machine Learning for Censorship Detection}
Our work is a machine learning-based approach for censorship detection. There are
studies that propose machine learning approaches for censorship
detection. For instance, a study uses machine learning to understand Chinese
social media censorship~\cite{li2015predicting}.  However, these approaches are
for the host-based or the server-based censorship because the data used in
their studies are the content of the blogs being censored.  Different from
those, our work is on network-based Internet censorship detection in that the
data on which our approaches rely are network measurement data and do not include
the online content being censored.

Additionally, machine learning has been used to design sophisticated
Internet censorship methods or to recognize certain types of network
traffic~\cite{gao2021machine, trivedi2016fully}. This trend adds urgency to studies
on machine-learning based censorship detection like this one. This necessity of this
work is increased given that the literature on detecting
network-based Internet censorship using machine learning is still rare.

\paragraph{Learning from Network Measurement Data}
This work is about learning from the network reachability data collected by
censorship measurement platforms to detect Internet censorship. The network
reachability data is a type of network measurement data. A deep learning model
has been used to characterize Internet hosts. In particular, a variational
autoencoder, an unsupervised neural network model, was previously designed to construct
low-dimensional embeddings of the high-dimensional binary representations of
the IP intelligence data collected by Censys, a type of network measurement
data~\cite{sarabi_characterizing_2018}.  Our work significantly
benefits from this and in spirit is an application of such an approach. 
However, there are
several differences.  First, the problem we investigate, network-based
Internet censorship detection, is distinct from host characterization. Second, our
approach to infer the low-dimensional embeddings is a sequence-to-sequence
model, which is motivated by the observation that the order of data elements
matters. Third, we compare the latent feature representation approach with a
state-of-the-art image-based approach. Finally, we design an engineering
solution to overcome the hurdle that the censorship network 
measurement data are enormous.

%% file: system.tex
\section{Censorship Feature Representation and Detection}
\label{sec:design}

\begin{figure*}[!htbp]
	\centering
	\begin{subfigure}[b]{\textwidth}
	\centering
	\begin{tikzpicture}
		\tikzset{
			node distance=0.2in and 0.2in,
			inout/.style={rectangle, draw, dashed, black, align=center, font=\footnotesize,
			minimum height=0.5in, rounded corners=0pt},
			box/.style={rectangle, draw, solid, black, drop shadow, fill=white, align=center, font=\footnotesize,
			minimum height=0.5in, rounded corners=3pt}
		}
		\node[inout](data){Network \\ Reachability \\ Data};
		\node[box, right=of data](tovec){Preprocessing \\ \& \\Vectorizing};
		\node[inout, right=of tovec](vector){Raw \\ Vector};
		\node[box, right=of vector](ae){Sequence-to-Sequence\\Autoencoder\\(Censor2Seq)};
		\node[inout, right=of ae](embedding){Embeddings};
		\node[box, right=of embedding](classifier){Censorship\\Classifier\\(CD Classifier)};
		\node[inout, right=of classifier](decision){Deicision \\(Censorship\\Probability)};

		\draw[-latex](data) -- (tovec);
		\draw[-latex](tovec) -- (vector);
		\draw[-latex](vector) -- (ae);
		\draw[-latex](ae) -- (embedding);
		\draw[-latex](embedding) -- (classifier);
		\draw[-latex](classifier) -- (decision);

	\end{tikzpicture}
	\caption{An End-to-End Network-based Internet Censorship Detection Pipeline with Latent Feature Representation Learning}
	\label{fig:pipeline}
	\end{subfigure}
	\vskip 1.0em
	\begin{subfigure}[b]{\textwidth}
	\centering
	\begin{tikzpicture}
		\tikzset{
			node distance=0.2in and 0.2in,
			inout/.style={rectangle, draw, dashed, black, align=center, font=\footnotesize,
			minimum height=0.5in, rounded corners=0pt},
			box/.style={rectangle, draw, solid, black, drop shadow, fill=white, align=center, font=\footnotesize,
			minimum height=0.5in, rounded corners=3pt}
		}
		\node[inout](data){Network \\ Reachability \\ Data};
		\node[box, right=of data](tovec){Preprocessing \\ \& \\Vectorizing};
		\node[inout, right=of tovec](vector){Raw \\ Vector};
		\node[box, right=of vector](imaging){Character-to-Pixel\\Processing};
		\node[inout, right=of imaging](images){Gray-Scale\\Images};
		\node[box, right=of images](classifier){Censorship\\Classifier\\(DenseNet)};
		\node[inout, right=of classifier](decision){Deicision \\(Censorship\\Probability)};

		\draw[-latex](data) -- (tovec);
		\draw[-latex](tovec) -- (vector);
		\draw[-latex](vector) -- (imaging);
		\draw[-latex](imaging) -- (images);
		\draw[-latex](images) -- (classifier);
		\draw[-latex](classifier) -- (decision);

	\end{tikzpicture}
	\caption{An Alternative Network-based Internet Censorship Detection Pipeline with Image Classification}
	\label{fig:pipeline:densenet}
	\end{subfigure}
	\caption{Processing pipelines of proposed models}
	\label{fig:models}
\end{figure*}
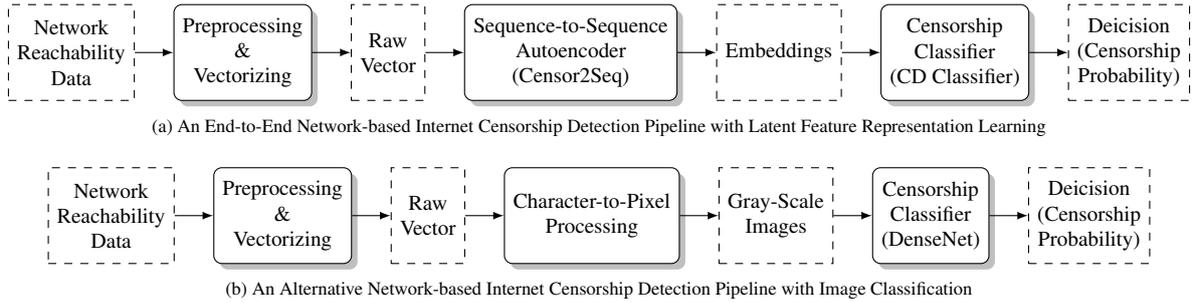

We propose an end-to-end network-based Internet censorship detection system 
whose processing pipeline is 
as shown in Figure~\ref{fig:pipeline}. The pipeline begins with converting
network reachability data to numerical vectors, i.e, raw vectors as detailed in
Section~\ref{sec:design:vector}.  We design a sequence-to-sequence autoencoder
model as the next processing component in the pipeline, and it is to infer
latent feature representation in embedding space (called embedding vectors)
from the raw vectors. We discuss this component in Section~\ref{sec:design:fl}.
Finally, we classify each embedding vector by a fully-connected neural
network, which we present in Section~\ref{sec:design:cd}. For
convenience, we refer to this system as the E2ECD model. To evaluate this
proposed system, we propose an alternate one as in
Figure~\ref{fig:pipeline:densenet} and document it in
Section~\ref{sec:design:alternate}.

\subsection{Structured and Textual Data to Numerical Vectors}
\label{sec:design:vector}

The censorship measurement platforms run tests to collect network reachability
data. A test typically consists of a request and a response where a no-response
or timeout event is considered a response. There are {\em three challenges} to process
the data collected by the measurement platforms.  First, the network
reachability data recorded by the censorship measurement platforms are
large and overwhelm our local storage capacity, and likely that of other
researchers who wish to pursue a similar research agenda.  Second, the requests
and responses are structured data. For instance, Censored Planet runs HyperQuack
tests where a test consists of an Echo request and a response, and records the
request and the response in the JSON format. Figure~\ref{fig:quack:data} is an
example data record. As shown, the entire data record is structured.
Furthermore, the values of some data fields are either structured or
semi-structured. For instance, the numbers in an IP address have different
meaning as the address has both host number and network number while the
response field can be in several formats, such as, in HTML with embedded
JavaScript (as illustrated in Listing~\ref{fig:quack:data}). 
Third, the combination of the overwhelming amount of data and the feature 
representation learning via deep neural networks can demand excessive computational resources, and thus there is need to balance feature representation learning effectiveness and required computational resources.

\begin{lstlisting}[float=tbh, style=jsonstyle, % 
caption={An example of HyperQuack test data record. We truncated the record to save
space. The response field is an array, and as such, an actual record can have multiple response bodies, where each response body corresponds to a block page; however, 
we transform each of those records to multiple records, each of which has a single 
response body. See Section~\ref{sec:eval:data} for detailed discussion.}, %
label=fig:quack:data]

{
  "vp": "114.116.151.108",
  "location": {
    "country_name": "China",
    "country_code": "CN"
  },
  "service": "echo",
  "test_url": "666games.net",
  "response": [
    {
      "matches_template": false,
      "response": "HTTP/1.1 403 Forbidden\nContent-Type: text/html; charset=utf-8\nServer: ADM/2.1.1\nConnection: close\nContent-Length: 520\n\n<html>\n\t<head>\n\t\t<meta http-equiv=\"Content-Type\" content=\"textml;charset=GB2312\" />\n\t\t<style>body{background-color:#FFFFFF}</style> \n\t\t<title>\u975e\u6cd5\u963b\u65ad222</title>\n\t\t<script language=\"javascript\" type=\"text/javascript\">\n         window.onload = function () { \n           document.getElementById(\"mainFrame\").src= \"http://114.115.192.246:9080/error.html\";\n            }\n\t\t</script>   \n\t</head>\n\t<body>\n\t\n\t\t<iframe style=\"width:100%; height:100%;\" id=\"mainFrame\" src=\"\" frameborder=\"0\" scrolling=\"no\"/>\n\t</body>\n</html>\n",
      "start_time": "2021-07-19T01:07:35.692415469-04:00",
      "end_time": "2021-07-19T01:07:37.135588863-04:00"
    },
  ],
  "anomaly": false,
  "controls_failed": false,
  "stateful_block": false,
  "tag": "2021-07-19T01:01:01"
}
\end{lstlisting}

\paragraph{Streaming I/O for Large Scale Machine Learning} To address the
overwhelming amount of data records, we opt for an engineering solution that can
iterate through the data directly from their cloud storage and process the data
records as a data stream. Our engineering solution based on
WebDataset~\cite{aizman_high_2019} is publicly available as a Python
module~\cite{shawn_p_duncan_censored_2022}.  

\paragraph{Vectorizing Structured Data}
Our aim here is to capture nuances from the structured network measurement data records, and at the same time produce relatively small input vectors, which is driven in part to address the latter two challenges discussed above. 

An example of the Censored Planet measurement data record is given in
	Listing~\ref{fig:quack:data}. Each record can have multiple responses for a
	test request value.  Because of this, to begin to process such test data, we
	pair the top level metadata corresponding to a request to each unique
	response to form a test record.  Such data records are structured as it
	follows a tree-like structure as dictated by its syntax, e.g., in JSON syntax
	and contains multiple data types of data.
	As required by machine learning models, we need to
	transform each test record to a numerical 
	vector~\cite{sarabi_characterizing_2018}.  It is important to note
	that we treat a test record as a sequence in order to preserve the
	structure of the field values.  We generally follow the
	spirit of the prior method that parses JSON data
	records~\cite{sarabi_characterizing_2018} to transform test records to
	sequences in a consistent and predictable manner, however 
	with several distinctions that we detail below.  The end result is a set of
	numerical vectors. For convenience, we
	refer to the numerical vector of a test record as the vector representation of
	the record or the raw vector. 

\begin{figure*}[!htbp]
	\centering
	\begin{tikzpicture}
		\tikzset{
			node distance=0.2in and 0.2in,
			inout/.style={rectangle, draw, dashed, black, align=center, font=\footnotesize,
			minimum height=0.25in, rounded corners=0pt},
			box/.style={rectangle, draw, solid, black, drop shadow, fill=white, align=center, font=\footnotesize,
			minimum height=0.25in, rounded corners=3pt}
		}
		\node[inout](url){\code{test\_url} field};
		\node[inout, below=of url](resp){\code{response} field};
		\node[box, right=of resp, yshift=0.25in](token){Extract text token\\using XML-R};
		\node[box, right=of token](dict){Form \\ token dictionary};
		\node[box, right=of dict](xmlr){Index text tokens  from\\pretrained XML-R};
		\node[box, right=of xmlr](filter){Filter out\\token indices\\per the dictionary};
		\node[inout, right=of filter](vector){Token index \\vectors};

		\draw[-latex](url) -- (token);
		\draw[-latex](resp) -- (token);
		\draw[-latex](token) -- (dict);
		\draw[-latex](dict) -- (xmlr);
		\draw[-latex](xmlr) -- (filter);
		\draw[-latex](filter) -- (vector);

	\end{tikzpicture}
	\caption{Multilingual data processing of the \code{test\_url} and \code{response} fields and dimensionality reduction via 
	a pretrained XLM-R model~\cite{conneau2020unsupervised}. 
	The \code{response} field can contain natural language text and program code. To reduce the size of the encoding vectors, We
	filter out those with low frequency, such as, those do not appear in our data (i.e., frequency is 0) or rarely appear depending
	on the constraint of the available computational resources. In this work, by filtering out tokens with frequency of 0, we reach
	sufficiently small embedding vectors of length of 6,813.}
	\label{fig:xlmr}	
	
\end{figure*}
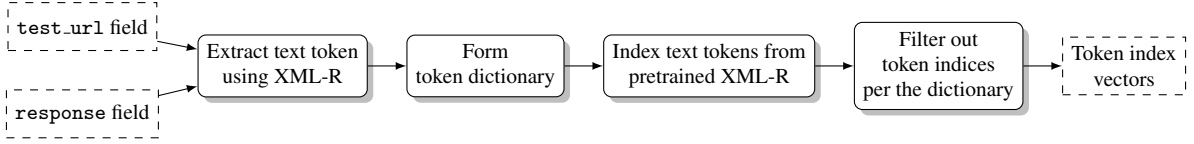

The transformation method is as follows. First, we process simple data fields,
e.g., \code{Datetime} fields in ISO format are transformed to UNIX timestamps, \code{Boolean}
fields transformed to a 1 or 0, and we split and convert the IP addresses into
sequential integers. We hold all this simple integer data and encode it as described below.
Second, two fields that contain text, \code{test\_url} and \code{response} need a more specialized
approach.  The \code{response}
field can contain the HTTP response code and
string, the HTTP headers, and the markup of a block page with text in a variety of
languages, e.g., the \code{response} field in the example in
Figure~\ref{fig:quack:data} is a concatenation of the response code and string concatenated, the
HTML headers, and HTML that contains an embedded
JavaScript. Some versions of Censored Planet data separate these elements into their own fields within
a nested JSON object.  We concatenate the \code{test\_url} followed by the \code{response} data into a single string. This string is relatively long when compared to the rest of the field. Aiming to reduce smaller input vectors, 
we then take advantage of the XLM-R multilingual transformer model
to process this string into numeric tokens~\cite{conneau2020unsupervised}. Figure~\ref{fig:xlmr} illustrates the processing pipeline
for these data fields. 

The XLM-R transformer uses just over 250,000 tokens. Given the amount of the data records we process, 
the demand of both memory and CPU time during feature representation learning is unbearable.  
We reduce the size of the token vocabulary so that
the training tasks would fit within the computing capacity available to us.  We do this by creating a
sequential index of the 6,813 XLM-R tokens needed to encode all the text in our datasets. We use the index
value in our vectors rather than the corresponding XLM-R value. The maximum value in our simple integer data
described above is 256.  We encode these integers as tokens also, so there are 7,069 possible token values
which is substantially smaller that XLM-R vocabulary.  The final raw vector begins with the variable length encoded
text which begins with XLM-R sequence start and sequence end tokens.  Appended to this variable length data are the simple integer tokens
which are also surrounded by XLM-R sequence start and sequence end tokens.

\subsection{Feature Representation Learning}
\label{sec:design:fl}
The next step in our method is to learn latent feature representations from the raw vectors.
For this we build an unsupervised
sequence-to-sequence neural network model. For convenience, we call this model
Censor2Seq.

The latent feature representations are often
referred to as embeddings~\cite{sarabi_characterizing_2018}. The learning here
is essentially to determine an embedding function $E : X \rightarrow Y$ that
takes an input $x \in X$ in a $m$-dimensional space and produces its
corresponding numerical vector $y \in Y$ in a n-dimensional vector space, where
$n \ll m$. The advantage of the embeddings are well discussed in the
literature~\cite{sarabi_characterizing_2018}. A primary advantage is that a
well-designed low-dimensional embedding function can infer embedding vectors
that retain information in raw input data in much higher dimension and offers
significant computational advantage due to low dimensionality and the 
continuity of the embedding space. 

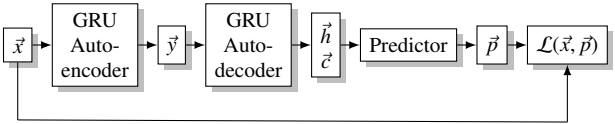
\begin{figure}[!htbp]
	\centering
	\begin{tikzpicture}
		\tikzset{node distance=0.3in and 0.1in,
			box/.style={rectangle, draw, drop shadow, fill=white, align=center, 
			font=\footnotesize},
		}
		\node[box](x){$\vec{x}$};
		\node[box, right=of x](enc){GRU\\Auto-\\encoder};
		\node[box, right=of enc](y){$\vec{y}$};
		\node[box, right=of y](dec){GRU\\Auto-\\decoder};
		\node[box, right=of dec](zc){$\vec{h}$\\$\vec{c}$};
		\node[box, right=of zc](pred){Predictor};
		\node[box, right=of pred](xcap){$\vec{p}$};

		\node[box, right=of xcap](loss){$\mathcal{L}(\vec{x}, \vec{p})$};

		\coordinate[below=of x](c1){};
		\coordinate[below=of loss](c2){};

		\draw[-latex](x) -- (enc);
		\draw[-latex](enc) -- (y);
		\draw[-latex](y) -- (dec);
		\draw[-latex](dec) -- (zc);
		\draw[-latex](zc) -- (pred);
		\draw[-latex](pred) -- (xcap);

		\draw(x) -- (c1);
		\draw(c1) -- (c2);
		\draw[-latex](c2) -- (loss);
		\draw[-latex](xcap) -- (loss);

	\end{tikzpicture}
	\caption{The architecture of the encoder-decoder model for
	sequence-to-sequence feature representation learning where $\vec{x}$ is
	a raw vector in a high-dimensional space, $\vec{p}$ is akin to
	an estimate of $\vec{x}$, $\vec{y}$ is the embedding vector in a
	low-dimensional embedding space, and $\mathcal{L}$ is the loss function. The
	learning algorithm is to minimize $\mathcal{L}$.}
	\label{fig:model:arch}
\end{figure}

We follow the typical encoder-decoder
architecture~\cite{sarabi_characterizing_2018, rezende2014stochastic} to design
the feature representation learning model. The architecture of the model is in
Figure~\ref{fig:model:arch}. 
The model takes the raw numerical
vectors as input. Each vector is a sequence of tokens, 
the set of tokens of all the row vectors is the ``vocabulary'', and each token is represented by an
integer that indexes the token in the vocabulary. The model consists of an
encoder block, a decoder block, and a predictor block. 

The predictor is straightforward. The inputs to the predictor are the
concatenated vector of a weighted context vector ($\vec{c}$) and the state
vector ($\vec{h}$) of the decoder block that we shall discuss later. 
The predictor outputs a series of probabilities ($\vec{p}$)
across the range of possible tokens, representing a discrete probability mass
function over the vocabulary. 
The weighted context vector here is to capture
the importance of different source-side information (e.g., which element in the state
vector) for predicting the probability of a particular token in the vocabulary~\cite{luong2015effective}.
We shall give a more detailed treatment of the context vector below (See the discussion about Equation~\eqref{eq:attention}). 
These probabilities are then used to compute a
cross-entropy loss at each step in the decoding process. The summation of
these losses for all decoding steps is used as the loss for the current input. 
The cross-entropy loss function is well-known and is estimated as follows, 

\begin{equation}\label{eq:loss}
	\mathcal{L}(\mathbf{x}, \mathbf{p}) = - \frac{1}{m}
		\sum\limits_{i=0}^{m-1}
			\sum\limits_{t=0}^{T-1}
				{p}_i({x_i}_{t})
\end{equation}

\noindent where $\mathbf{x}$ represents the batch of training data consisting
of $m$ raw vectors of censorship test records, $\vec{x_i}$ the $i$-th vector
of $\mathbf{x}$, ${x_i}_{t}$ the $t$-th item of $\vec{x_i}$,
${p}_i(\vec{x_i}_t)$ its probability mass, and $T$ the length of
$\vec{x}$ or the total time steps.

The encoder is a bidirectional Gated Recurrent Unit (GRU). It takes a vector
representation as input and outputs a sequence of states, i.e., $h_t$ at time
step $t$. The final state
summarizes the entire vector representation and is the embedding of the test
record, i.e., $\vec{y}$ in Figure~\ref{fig:model:arch}. Note that 
$\vec{y} = h_{T-1}$  where $T-1$ is the last temp step.

The complexity of the model is in the decoder where we take advantage of
several mechanisms to help with training speed and convergence. The decoder
takes the output of the encoder (i.e., $\vec{y}$) as input and attempts to
restore the input to its original form (i.e., $\vec{x}$). 

First, before processing an input sequence, we use a simplification of
scheduled sampling~\cite{raff_inside_2021, mihaylova-martins-2019-scheduled}.
This sampling selects the inputs for the decoder by randomly switching
between \textit{auto-regression} and \textit{teacher forcing}. 
The difference between these two strategies
begins with the second item of the input sequence (i.e., $\vec{y}$). 
If auto-regression has been
chosen the inputs will be sampled from the range of the vocabulary using the
probabilities produced by the predictor block (i.e., $\vec{p}$).  If teacher
forcing is chosen, the decoder inputs are the same as the inputs to the
encoder. 

The decoder is a sequence of bidirectional Gated Recurrent Unit Cells. We enhance
it with an attention mechanism~\cite{niu2021review}, in particular, the ``dot'' 
attention mechanism~\cite{luong2015effective}. 
Instead of using fixed
neuron weights, the attention mechanism allows us to enhance the 
effect of some parts of the input data while diminishing other parts for predicting
token probabilities via attention scores. 
In one
direction, each token in the input sequence is processed through the GRU cells.
In the other direction, we calculate attention scores. Attention scores are
computed using a dot product~\cite{raff_inside_2021}

\begin{equation} \label{eq:attention}
score(h_t, c_t) = \frac{h_t^T \cdot c_t}{\sqrt{H}}
\end{equation}

\noindent with values of $h_t$ taken from the state produced at step $t$ by the
encoder, $h_t^T$ is $h_t$'s transpose, $c_t$ is the context, 
and $H$ is the
input size used for the GRU Cell. We compute $c_t$ as the decoder output at step $t$.
We use the attention scores to weight the input sequence at each step of the decoding process.

The use of the model consists of two phases, training and
inference.  Given a set of vector representation of the test records, we train
the model by minimizing the entropy loss function in equation~\eqref{eq:loss}.
With a trained model, we infer the embedding of a test record by feeding the
model with the numerical vector representation of the test record 
and take the output of the
encoder as the embedding.  The model is an unsupervised learning model during
training and inference, and as such, no labels of censorship are given.

\subsection{Censorship Classification}
\label{sec:design:cd}
An advantage of embedding is its versatility, due in part to the continuity
of the embedding space. It can be fed to a variety of
machine learning models. The central question we want to answer is,
``is the reachability problem the result of censorship?''  
For this purpose, we construct a multiple-layer fully-connected 
dense neural network with three layers as a binary classifier: an input layer
matching the dimension of the embeddings produced by the autoencoder, an output
layer that further refines the output of the hidden layer to a probability, and
a hidden layer in between. For convenience, we call this neural network
the CD classifier.

\subsection{Alternative Approach}
\label{sec:design:alternate}

Prior works typically adopt a semi-automated approach for Internet censorship 
detection, where they first leverage an unsupervised machine learning algorithm,
such as, a clustering algorithm to divide Internet outage measurements, e.g.,
block pages into clusters, and second, manually examine the clusters to determine
whether the cluster represents a type of Internet 
censorship~\cite{raman_measuring_2020, niaki2020iclab}. A necessary step
in the process is to extract data features from the internet
measurement data. Page length and term frequency vectors are two features
borrowed from natural language processing to cluster block 
pages~\cite{jones2014automated}. Although the clustering algorithm
using the two features reduces manual efforts to examine the clusters,
several follow-up studies indicate that the features can lead to 
high false positives due to natural variations in page length from dynamic
and language-specific content~\cite{yadav2018light, raman_measuring_2020}. 
Recognizing the limitation of the two features, researchers are seeking 
different feature representation methods for Internet measurement data, such as,
block pages~\cite{raman_measuring_2020}.

Deep learning has found tremendous success in computer
vision~\cite{huang_densely_2017, li_fei-fei_imagenet_2012}. In contrast to 
more traditional learning algorithms, such as, Super-Vector Machine and 
Tree-based algorithms (Decision Tree, Random Forest) for which we need to
extract from the raw data manually designed features to represent the data
in a tabular form, deep learning can learn to extract features directly
from the raw data, which, numerous studies show it is an advantage because
we do not always know what the best features might be for a particular 
machine learning application~\cite{bengio2013representation}. 
A clever way to
design a learning system for non-tabular data is to take
advantage of these recent advances in deep learning by encoding the data as
gray-scale images and use a deep learning model for computer vision that is
readily available to learn
from these images. There have been applications of this approach in several
domains, such as, for malware detection~\cite{hemalatha_efficient_2021} and for
software defect prediction~\cite{chen2020software}.  A gray-scale image can be
encoded using a byte pallet, in which the range from black to white is encoded
as an integer between 0 and 255. In other words, we can consider a gray-scale
image as a vector of $W \times H$ integers where $W$ is the width of the image
and $H$ the height. For our structured data, after converting a
censorship test record similar to that in Listing~1 into a
raw vector, we pad the
beginning and end of the vector with sufficient 0 to 
equal a uniform length of size $W \times H$. We then export this vector to a
sequence of bytes, reshape the vector to a $W \times H$ array from which we
form the gray-scale image.

Recently several works
have adopted this approach in their semi-automated censorship detection
pipeline and find that the approach leads to improved predictive performance
than those using the page length and term frequency vector 
features~\cite{raman_measuring_2020, sundara_raman_censored_2020}. 
Following these prior works, we build a fully-automated Internet censorship detection model
where we use images as feature representations of block pages. For this,
we elect to use the well-known DenseNet
model~\cite{huang_densely_2017}.  DenseNet uses the architecture of a Convolutional
Neural Network (CNN). A DenseNet has multiple CNN blocks that have the
following neuron connectivity pattern: each layer in the CNN block
is connected to all the others with the block. Due to this connectivity
pattern, there is a shortcut path between each individual layer to the loss
function and the training of the layer can be directly supervised by the loss
function. Aside from this, the connectivity pattern also results in a compact
network that is less prone to overfitting and encourages heavy feature reuse.
Because of this, prior work suggests that DenseNet should be a natural fit for per-pixel
prediction problems~\cite{zhu2017densenet}.

DenseNet is available as a pre-trained model~\cite{pytorchdensenet} 
that expects images sized $224 \times 224$. The pre-trained model has a final linear
layer of size 1,000 as a classifier for 1,000 image categories.  We replace
that linear layer in the pre-trained model with a new layer accepting the same
input dimension but producing a single output, the probability at which 
the input is a censorship event.  

As an alternative to for the proposed End-to-End
Censorship Detection model (E2ECD), we consider a modeling pipeline as
illustrated in Figure~\ref{fig:pipeline:densenet}. For convenience, we refer to
this model as DenseNetCD. Because the alternative model leverages the feature presentation method
that prior works consider the best and is equipped with an identical classifier as the
E2ECD model, we shall use it as a baseline model to compare
with the E2ECD model.

%% file: eval.tex
\section{Evaluation}
\label{sec:eval}

\subsection{Evaluation Data Sets}
\label{sec:eval:data}

We prepare two data sets by extracting Quack test records from Censored 
Plant~\cite{sundara_raman_censored_2020}. One data set is for training and
evaluating the latent feature learning model while the other for training and
evaluating censorship detection.  For convenience, we refer to the
former data set as Data Set FL while the later Data Set CD. 

It is important to note that Censored Planet organizes their data based on
tests it runs. Specifically, each test is defined by its test request sent
from a vantage point, and this test is to run multiple times, which results in
multiple responses for the same test request message. On Censored Planet, a
record consists of the metadata about the test request message and the
multiple response messages. We flatten each Censored Planet test record by
iteratively pairing the top level request metadata with each response message
to create a record, essentially a record akin to a single request-response pair. 
This is to
match the intended use of the system, i.e., for each request sent, we want to
determine whether the request is censored based on the response received. In
the remainder of this article, a test record is like a request-response pair as described,
rather than a test record on Censored Planet.

Data Set FL
consists of \FLnrecords response records of the tests that Censored Plant ran
while Data Set CD consists of \CDnrecords vectors created as described in Section \ref{sec:design:vector} paired with metadata. The metadata consists of the domain or string under test, the IP address and country of the vantage point, timestamp for the test, and a flag value for censored, uncensored or undetermined.

Censored Planet matches each test response with a censorship fingerprint
written in regular expression~\cite{sarah_laplante_blockpage_2021} and assigns
a label to each test record. The label can be either ``censored'', ``uncensored'',
or ``undetermined''. It is worth noting that the latent feature representation
learning model in Section~\ref{sec:design:fl} is an unsupervised learning
model, as such, the label is irrelevant and unused. However, the censorship
classification model in Sections~\ref{sec:design:cd}
and~\ref{sec:design:alternate} are supervised models and the labels are
necessary for training and also serve as the ``ground truth'' for evaluation.

Table~\ref{tab:data} summarizes useful statistics for the two data sets. 
\begin{table}[!htbp]
	\centering
	\caption{Evaluation Data Sets}
	\label{tab:data}

	\begin{tabular}{p{0.67\columnwidth} r}
		\toprule
		\multicolumn{2}{c}{Data Set FL (for Feature Learning)} \\
		\midrule
		\# of Training Records & \FLntrain \\
		\# of Validation Records & \FLnvalid \\
		Total  & \FLnrecords \\
		\midrule
		\multicolumn{2}{c}{Data Set CD (for Censorship Detection)} \\
		\midrule
		\# of Censored Records & \CDncensored \\
		\# of Uncensored Records & \CDnuncensored \\
		\# of Undetermined Records & \CDnunknown \\
		Total & \CDnrecords \\
		\bottomrule
	\end{tabular}
\end{table}

\subsection{Baseline Model}
\label{sec:eval:dense}
\begin{figure}[!htbp]
	\centering
	\begin{subfigure}[b]{0.8\columnwidth}
		\centering
		\includegraphics[width=\textwidth]{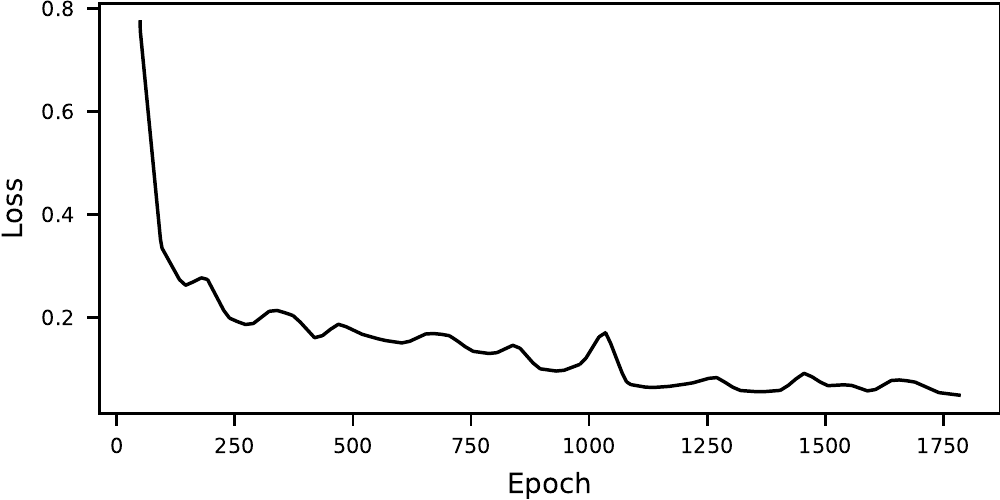}
		\caption{Validation Loss}
		\label{fig:dn_val_loss}
	\end{subfigure}
	\hfill
	\begin{subfigure}[b]{0.8\columnwidth}
		\centering
		\includegraphics[width=\textwidth]{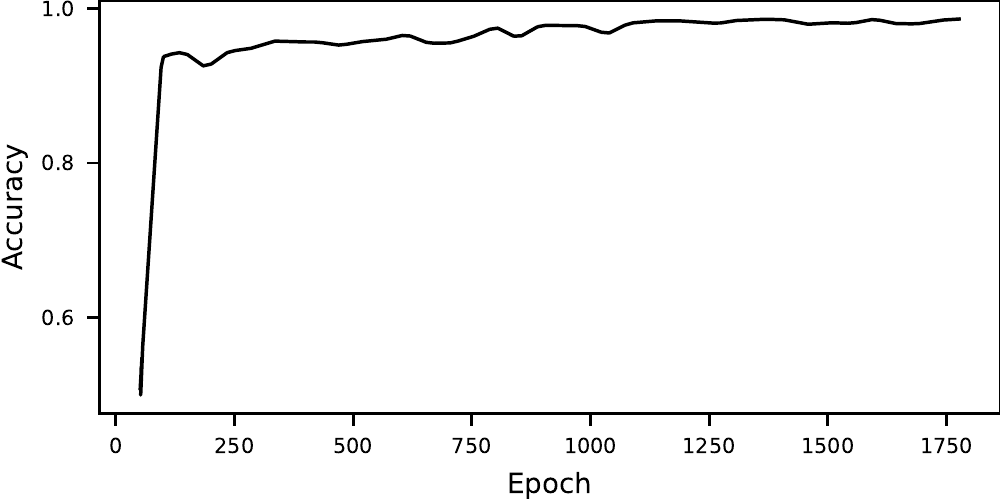}
		\caption{Validation Accuracy}
		\label{fig:dn_val_acc}
	\end{subfigure}
	\caption{Validation loss and accuracy over training epochs}
	\label{fig:dn:train}
\end{figure}

We treat the alternative model in Section~\ref{sec:design:alternate} as 
a baseline mode, and discuss its evaluation first here. 
We expand the vector in each record in Data Set CD to an vector of length $224
\times 224$ by padding each end
of the byte sequence with zeros if necessary, export the vector to bytes
and treat each vector as a gray-scale image. 
The training and evaluating settings are similar to those for the 
CD Classifier discussed in Section~\ref{sec:eval:cd:train} later.  
These include the hold-out evaluation setting, i.e., we randomly partition
the data set into three partitions, \CDtrainratio of the data for training,
\CDvalidratio for validation, and \CDtestratio for testing. In addition, we use a learning
rate scheduler that varies the learning rate between $1 \times 10^{-6}$ and $1
\times 10^{-7}$. The scheduler is  a cosine annealing 
strategy~\cite{loshchilov10sgdr} with period of
10 epochs. We monitor the training processing using the validation loss and
accuracy as shown in
Figure \ref{fig:dn_val_loss} and Figure \ref{fig:dn_val_acc}. The trained
network resulted in a loss of 0.048 on the test data with an accuracy of 0.986
where the accuracy is defined $(TP + TN)/(P + N)$.

\subsection{Feature Representation Learning}
The Censor2Seq model described in Section~\ref{sec:design:fl} is to learn
latent feature representation.  Table~\ref{tab:fl:params} summarizes the
hyperparameters of the Censor2Seq model used in this evaluation.

To evaluate the latent feature representation learning model, we use Data Set
FL that is divided into a training and a validation data set (Table~\ref{tab:data}).  
Following the procedure in Section~\ref{sec:design:vector}, we convert
each test record in the data set to a sequence of numerical values
represented as a numerical vector. This process is described in Section~\ref{sec:design:vector}
in which we detail how the largest token value in our vectors is 7,069. This value is used to define
the embedding range.

For embedding size, we experiment with a
number of values with a consideration on memory requirement, processing time,
and convergence, find that the acceptable values range from \FLembsizemin
to \FLembsizemax, and settle at size \FLembsize.

\begin{table}[h]
	\centering
  \caption{Hyperparameters of Censor2Seq Model}
  \label{tab:fl:params}
	\begin{tabular}{p{0.67\columnwidth} r}
    \toprule
		Hyperparameter & Value \\
		\midrule
		size of input sequence size & \FLseqlen \\
    embedding dimension & \FLembsize \\
		\midrule
    encoder hidden layer size & \FLenchidden \\
    encoder output layer size & \FLencoutput \\
		\midrule
    decoder input layer size & \FLdecinput \\
    decoder hidden layer size & \FLdechidden \\
		\midrule
		decoder output sequence size & \FLdecoutput \\
    \bottomrule
  \end{tabular}
\end{table}

As shown in Table~\ref{tab:data}, we partition the data set into two subsets,
one for training and the other for validation. To reduce the demand of
computational resources, such as, memory and CPU time, we adopt a statistical
gradient descent (SGD) mini-batch training approach where we randomly sample
\FLsamplingratio of training data, i.e., sample from \FLntrain records 
at each training
epoch and set batch size as \FLbatchsize. Following the best practice, we
select a simple learning rate scheduler that uses 
a cosine annealing strategy~\cite{loshchilov10sgdr}
that decays learning rate smoothly from \FLlrbegin to \FLlrend with period of
10 epochs.

\begin{figure}[!htbp]
    \centering
    \includegraphics[width=0.8\columnwidth]{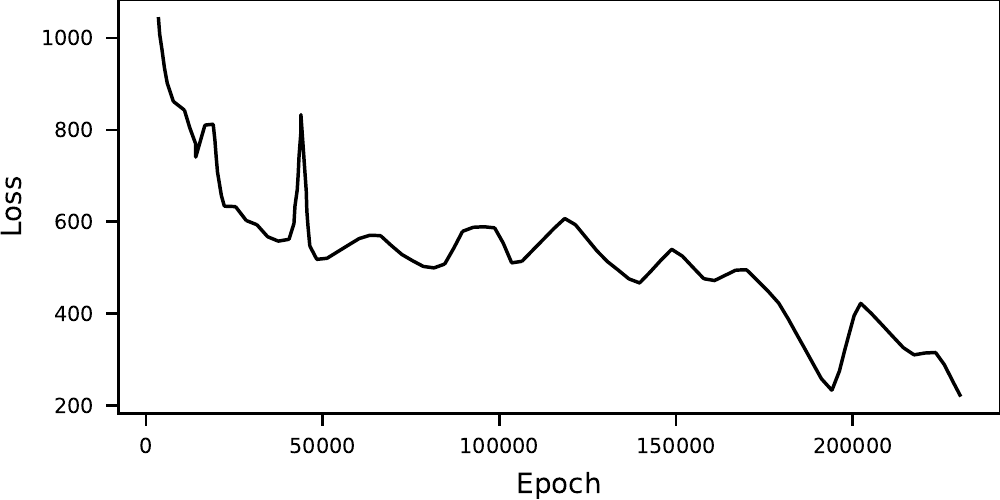}
    \caption{Validation loss of the Censor2Seq model.}
    \label{fig:ae_val_loss}
\end{figure}

We monitor the training process by checking on the entropy loss function value
computed on the validation subset following each epoch. Following the selected
early exit strategy, we stop the training process when the validation entry
loss function value twice dipped below \FLlossthreshold of the initial entropy
loss. The validation loss function value over training epochs is in
Figure~\ref{fig:ae_val_loss}.

The effectiveness of the Censor2Seq model is ultimately  reflected by the quality of
the lightweight numerical embeddings it produces.  In the Section~\ref{sec:eval:cd}, we evaluate how
effective the embeddings are to detect censorship events. 

\subsection{Censorship Detection}
\label{sec:eval:cd}

Section~\ref{sec:design:cd} describes the classifier for censorship detection.
It is a fully connected dense neural network. Table~\ref{tab:cd:params} is
the hyperparameters of the neural network. 
This neural network is a supervised classification model that requires
training.

\begin{table}[h]
	\centering
  \caption{Hyperparameter of Censorship Detection Neural Network}
  \label{tab:cd:params}
  \begin{tabular}{p{0.67\columnwidth} r}
    \toprule
    Parameter & \\
    \midrule
    input size & \CDinputsize \\ 
    first layer size & \CDfirstsize \\
    second layer size & \CDsecondsize \\
		output layer size & \CDoutputsize\\
    \bottomrule
  \end{tabular}
\end{table}

\subsubsection{Training}
\label{sec:eval:cd:train}

We use Data Set CD to train the CD classifier. For each test record in Data Set
CD, we process it using the trained Censor2Seq model. As a result, we obtain an
embedding vector for each test record in Data Set CD. We adopt a
hold-out evaluation strategy where we randomly partition the data set
into three partitions, \CDtrainratio of the data for training, \CDvalidratio
for validation, and \CDtestratio for testing. 

As shown in Table~\ref{tab:data}, the censorship data here is unbalanced with
regard to its label where there are 3 times as many responses identified as
uncensored as those censored.  We applied an approach of under-sampling the
majority class to balance the training dataset whenever labeled data are
needed, i.e., during the training we randomly under-sample the training data
from the uncensored records, the majority class to match the censored records,
the minority class.

\begin{figure}[!htbp]
  \centering
	\begin{subfigure}[b]{0.8\columnwidth}
		\centering
		\includegraphics[width=\textwidth]{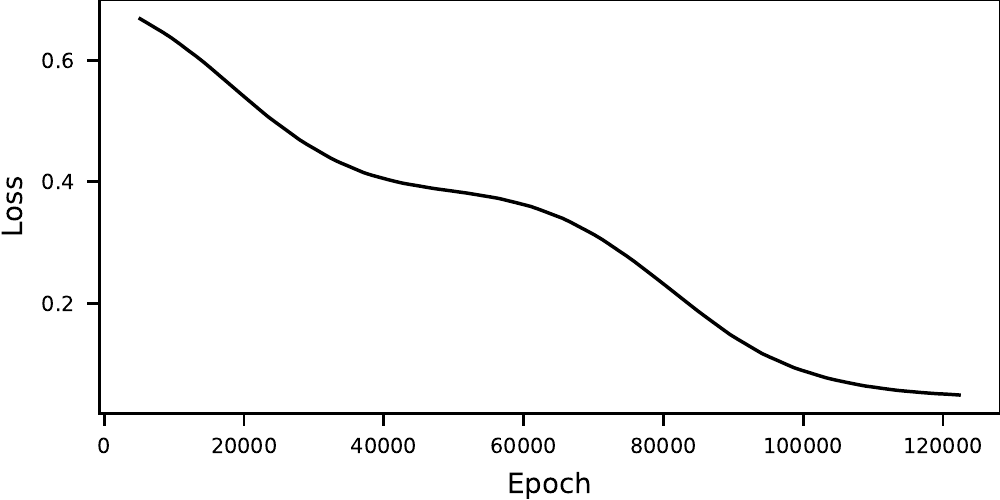}
		\caption{Validation Loss}
    \label{fig:le_val_loss}
	\end{subfigure}
	\hfill
	\begin{subfigure}[b]{0.8\columnwidth}
		\centering
		\includegraphics[width=\textwidth]{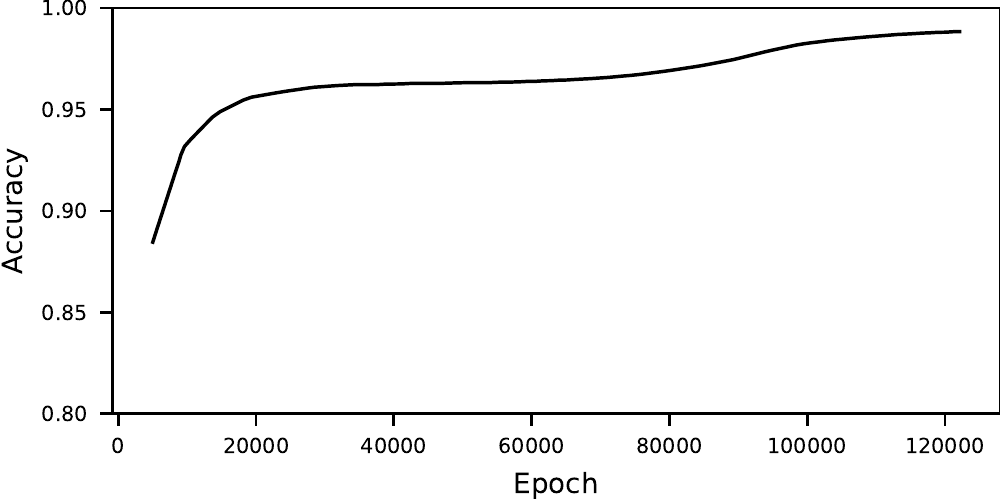}
		\caption{Validation Accuracy}
    \label{fig:le_val_acc}
	\end{subfigure}
	\caption{Validation loss and accuracy of the CD classifier over training epoch}
	\label{fig:cd:train}
\end{figure}

The learning rate scheduler uses a cosine annealing
strategy~\cite{loshchilov10sgdr} with the period of 10 epochs and varies the
learning rate between $1 \times 10^{-6}$ and $1 \times 10^{-7}$.  To prevent
overfitting, we check the entropy loss value computed on the validation data
after each training epoch and stop training when the validation loss ceases to
reduce. We aid this by using an early stopping strategy to eliminate
unnecessary excessive training time.  Figures \ref{fig:le_val_loss} and
\ref{fig:le_val_acc} illustrate the validation loss and the classification
accuracy over training epochs.

\subsubsection{Testing}
Finally, to evaluate the effectiveness of the classifier, a surrogate to the
effectiveness of the embeddings produced by the trained Censor2Seq model, we
classify the test data set with the trained classifier. The classifier results
in a loss of $0.047$ and an accuracy of $0.988$ where the accuracy is
defined $(TP + TN)/(P + N)$.

\begin{table}[!htbp]
	\centering
	\caption{Comparison of Classification Accuracy of E2ECD and DenseNetCD}
	\label{tab:eval:compare}
	\begin{tabular}{p{0.3\columnwidth} r r}
		\toprule
		Model & Loss & Accuracy \\
		\midrule
		E2ECD& 0.047 & 0.988 \\
		DenseNetCD & 0.048 & 0.986 \\
		\bottomrule
	\end{tabular}
\end{table}

Table~\ref{tab:eval:compare} summarizes and compares the evaluation results of
this model and the baseline model. The classification performance of the two
models are close if not considered identical. However, the E2ECD
model has a clear advantage when it comes to classifying
undetermined test records as discussed in Section~\ref{sec:eval:undermine}.

\subsection{Censorship Detection on Undetermined Records}\label{sec:results}
\label{sec:eval:undermine}

Data Set CD contains a \CDnunknown test records labeled as undermined as discussed
in Section~\ref{sec:eval:data}. 
We are interested to know whether there are censorship
events in this set of undetermined test records. For this, we classify the
records using both the trained CD Classifier and
DenseNetCD models. There is a striking difference between the results
of the models as shown in Table~\ref{tab:eval:undetermine}. To interpret the 
result, it is necessary to understand the measurement tests performed by 
Centered Planet. 

\paragraph{Background} The data records in Data Set CD are from Quack 
tests~\cite{vandersloot2018quack}. Two major phases of a Quack test are 
``Retry'' and ``Control''. In the ``Retry'' phase, the platform has
an Echo request with a potentially offending payload repeatedly sent by 
a vantage point, e.g., sent 5 times with a delay in between. When all 
requests result in a failure, the platform carries out the Control phase
where the platform has an Echo request with an innocuous payload sent by
the vantage point. Only when this Echo request fails, the test declares
that there might be a network reachability problem due to network 
interference. 

Quack is effectively detecting the network reachability resulted from network
interference~\cite{vandersloot2018quack}. However, not all of the network 
interference detected by Quack is censorship, The exceptions include server-side
blocking errors (e.g., HTTP Status Code 403), page-not-found errors (e.g., HTTP
Status Code 404), and DDOS checks on some domains~\cite{raman_measuring_2020}, In 
addition, a recent investigation also indicates that some Censored Planet test 
requests were not sent successfully, which could also lead to unexpected responses~\cite{niaki2020iclab}.

\begin{table}[!htbp]
	\centering
	\caption{Comparison of Classification Results on Undetermined Tests}
	\label{tab:eval:undetermine}
	\begin{tabular}{p{0.3\columnwidth} r r}
		\toprule
		Model & Probable Censorship & Probable\% \\
		\midrule
		E2ECD & 557,492 & 66.14\%\\
		DenseNetCD & 183 & 0.02\%\\
		\bottomrule
	\end{tabular}
\end{table}

\paragraph{Analysis of the Results}
Which model is more accurate? Our answer is not definite; however, it is more
likely that the CD Classifier produces more meaningful results, in particular, 
when considering the design of Quack, we expect a significant proportion of 
the undetermined records should be the results of censorship. Aiming to confirm
this, we divided each of the results into
groups based on the patterns of the data records.  Table~\ref{tab:ae_results} 
summarizes the grouping of the
classification result of the E2ECD model while Table~\ref{tab:dn_results} that
of the DenseNetCD model.

The E2ECD model via the CD Classifier yields far more censorship detections. We
can confirm with a high confidence that some of these results are indeed 
probable censorship events. For instance, groups 1--3 and 5--7 in 
Table~\ref{tab:ae_results} are the block pages of Internet filters (or Firewalls), 
such as, Bitdefender, Cisco Meraki, and Seqrite Endpoint Security, and 
in total, the 1,184 responses that are likely actual
censorship block pages by examining the patterns of the these block
pages. Similarly,  the 122 responses in group 4 that are an \textit{HTTP Code 302} response
are also likely the result of censorship 

Most of the remaining records are likely the result of censorship.  
An investigation on the
Great Firewall of China indicates that both \textit{connection reset by peer}
and \textit{connection timed out} are among its observable
characteristics~\cite{shu_data_2014}.  As discussed in the above, 
the response to the request sent to
these vantage points have been verified with innocuous payloads, and therefore
it is very unlikely that the network has unexpectedly and repeatedly failed
during these specific queries.

Table~\ref{tab:dn_results} summarizes the grouping of the result of the
DenseNetCD model.  There are far fewer results shown and all of them are block
pages or \textit{HTTP Code 302}. Clearly the ability of the DenseNetCD model
to discover new censorship events is questionable. 

We hypothesize that the difference between the models lie at the distinction
of the designs of the two approaches. DenseNet is designed for per-pixel prediction models.
As indicated in a recent study~\cite{cheng2021per}, the state-of-the-art per-pixel 
classification model underperforms mask classification models in image semantic 
segmentation tasks where the
mask classification models disentangle the image partitioning and classification
aspects of image segmentation. There is a parallel when it comes to the two censorship
detection models we proposed. The E2ECD model separates the censorship detection
into two tasks, feature representation learning and censorship detection while the DenseNetCD 
model bundles the two together. Additionally, the encoding method used by Censorship2Seq is also designed
to encode the relationships between the individual tokens while an image classifier is intended
to be aware of clusters of pixels, it is less aware of connections between distant pixels. Due to the design of the feature 
representation learning model and the disentanglement of the two tasks, E2ECD fares better
than DenseNetCD, as a result of more generalizable feature presentation learned by Censorship2Seq. It
is also highly likely that the DenseNetCD model learned highly idiosyncratic features specific to 
the training data, resulting in poor generalizability, and notably the
phenomenon of this nature has also been reported in studies using deep
learning models in other areas, such as, vulnerability prediction~\cite{chakraborty2021deep}. 

\begin{table*}
	\centering
  \caption{Censorship Candidates from E2DCD}
  \label{tab:ae_results}
	{\footnotesize
  \begin{tabular}{r p{0.25\textwidth} r p{0.55\textwidth}}
    \toprule
		No. & Type & Frequency & Locations\\
    \midrule
		1. & ATT block page & 3 & United States \\
		2. & Bitdefender Alert Page block page & 1,133 & India, United States \\
		3. & Extra Safe Internet block page & 4 & Netherlands \\
		4. & HTTP code 302 & 122 & Belarus, China, Kazakhstan, Pakistan, Russia \\
		5. & Meraki block page & 1 & Singapore \\
		6. & Seqrite Endpoint Security block page & 37 & India \\
		7. & Net Protector block page & 6 & India \\
		8. & `Connection: close` message returned without block page & 10 & China, U.S. Virgin Islands, United States \\
		9. & `connection reset by peer' error & 542,083 & Bangladesh, Brunei, Canada, China, Colombia, Egypt, Ethiopia, Germany, Hong Kong, India, Iran, Ivory Coast, Kenya, Libya, Mexico, Oman, Pakistan, Poland, Qatar, Russia, Saudi Arabia, Singapore, South Korea, Syria, Taiwan, Turkey, United Arab Emirates, United States, Uzbekistan, Vietnam \\
		10. & `no route to host' error & 296 & Argentina, Australia, Bangladesh, Belgium, Brazil, Colombia, Czechia, Germany, Greece, India, Indonesia, Italy, Japan, Libya, Malaysia, Mexico, Moldova, Nepal, Poland, South Korea, Spain, Switzerland, Taiwan, Thailand, United Kingdom, United States \\
		11. & `connection timed out' error & 2,885 & Angola, Argentina, Austria, Bangladesh, Brazil, Burundi, Canada, China, Colombia, Czechia, Germany, Greece, Guadeloupe, India, Iran, Italy, Japan, Mexico, Netherlands, Nigeria, Philippines, Russia, South Africa, South Korea, Spain, Sweden, Switzerland, Taiwan, Turkey, U.S. Virgin Islands, Ukraine, United Kingdom, United States \\
		12. & `I/O timeout' error & 6,983 & Algeria, Armenia, Bangladesh, Belarus, Belgium, Brazil, Burundi, Canada, China, Egypt, Germany, Guam, Hong Kong, India, Indonesia, Iran, Kazakhstan, Kenya, Madagascar, Malaysia, Mexico, Netherlands, Nigeria, Oman, Pakistan, Paraguay, Qatar, Romania, Russia, Saudi Arabia, South Korea, Taiwan, Tanzania, Thailand, Turkey, United Kingdom, United States, Uzbekistan, Vietnam, Zambia \\
		13. & `network is unreachable' error & 18  & Indonesia, Poland \\
		14. & `connection refused' error & 3,007 & Bangladesh, Brazil, Chile, China, Colombia, Czechia, Egypt, France, Germany, Greece, India, Italy, Ivory Coast, Japan, Kenya, Mexico, Nepal, Pakistan, Peru, Poland, Russia, Singapore, South Africa, South Korea, Spain, Sri Lanka, Sweden, Taiwan, Thailand, Ukraine, United States \\
		15. & `echo response does not match echo request' and no other data & 904 & Algeria, Armenia, Bangladesh, Belarus, Belgium, Brazil, Burundi, Canada, China, Egypt, Germany, Guam, Hong Kong, India, Indonesia, Iran, Kazakhstan, Kenya, Madagascar, Malaysia, Mexico, Netherlands, Nigeria, Oman, Pakistan, Paraguay, Qatar, Romania, Russia, Saudi Arabia, South Korea, Taiwan, Tanzania, Thailand, Turkey, United Kingdom, United States, Uzbekistan, Vietnam, Zambia \\
  \bottomrule
\end{tabular}
	}
\end{table*}

\begin{table*}
	\centering
  \caption{Censorship Candidates from DenseNetCD}
  \label{tab:dn_results}
	{\footnotesize
	\begin{tabular}{r p{0.25\textwidth} r p{0.55\textwidth}}
    \toprule
		No. & Type & Frequency & Locations\\
    \midrule
		1. & ATT block page & 3 & United States \\
		2. & Bitdefender Alert Page block page & 116 & India, United States \\
		3. & Extra Safe Internet block page & 1 & Netherlands \\
		4. & HTTP code 302 & 62 & China, Kazakhstan, Pakistan, Russia \\
		5. & Meraki block page & 1 & Singapore \\
  \bottomrule
\end{tabular}
		}
\end{table*}

%% file: threats.tex
\section{Limitations and Threats to Validity}
\label{sec:threats}

This work is on the design and the evaluation of an ``end-to-end''
network-based Internet censorship detection using deep learning. In particular,
we advocate and argue that there is a strength in the approach that uses latent
feature represent learning in embedding space. Deep learning is not clairvoyant
and there can be threats to validity and limitations that constrain the
potential of this work.  Here, we discuss several of these limitations.

\paragraph{Data}
We selected Censored Planet as the source of the evaluation data and extracted
Quack test data from it.  Because Quack~\cite{vandersloot2018quack,
raman_measuring_2020} monitors network side-channels, as we discussed in
Section~\ref{sec:related}, the censorship events detected are unambiguously
network-based Internet censorship.  

There is a threat to internal validity stemmed from the quality of the
evaluation data. A recent study indicates that a significant portion of the
Quack tests did not successfully transmit a test request in the first
place and resulted in false block pages~\cite{niaki2020iclab}. 

However, this impacts our results little. Since these false block pages are
not to be identified as a censorship and are invalid pages, there is no impact on the quality of the
test records that are labeled as censorship and our trained models can 
recognize almost all the correctly identified censorship events as shown
in Table~\ref{tab:eval:compare}.

\paragraph{Hyperparameter and Model Tuning}
The predictive performance of the deep learning models like those investigated
in this study is subject to the choice of
hyperparameters~\cite{yang2020hyperparameter}. Due to the computation resources
available to us, we were unable to experiment exhaustively on a large
collection of hyperparameters. However, we took measures to ensure the evaluation 
settings of the proposed
E2ECD and the DenseNetCD models are comparable. First, we adopted an identical
training strategy, i.e., the Stochastic Gradient Descent with the cosine
annealing learning rate scheduler with similar configuration. Second, we use
an identical hold-out evaluation setting. Finally, we identify a pre-trained
DenseNet that represents the state-of-the-art to design the baseline 
model. The proposed E2ECD model yields an almost identical prediction
accuracy as the baseline model 
on the test data set (Table~\ref{tab:eval:compare}). We are convinced that the latent
feature presentation model presented here has an appropriate design. 

Despite this, we recognize that we used the smallest of the DenseNet
architectures, also known as DenseNet 121 due to the constraint of
computational resources available to us.  The original DenseNet research found
that larger DenseNets had a lower error rate on the ImageNet classification
problem~\cite{huang_densely_2017}, as such, larger DenseNets may also be more
effective at detecting censorship. 

\paragraph{Censorship Detection on Undetermined Test Records}
To compare the E2ECD and the DenseNetCD models, we also test them on a set of
undetermined test records. The E2ECD model detects far more candidate
censorship events than the DenseNetCD model. We argue that it is likely the
result of the advantage of the latent feature representation approach.  The
undetermined test records are all perceived network outages.  First, it is
clear that we are unable to access the intention of any factors that might
cause these outages. Second, we are limited by our knowledge to determine
whether the outages are truly censorship or ordinary network outages. As the
result, the reliability of our assessment is not absolute.

Because the evaluation data set contains in total 842,842 records, a thorough
examination of the prediction results manually is infeasible. We plan to study
these test records in a future work. Because we are unable to access people's intention
who might initiate the censorship events, the correctness of censorship 
detection is often via cross-examination and inference. For this, we plan to
compare records collected by multiple Internet censorship measurement 
platforms including both the Censored Planet and 
ICLab in the same time period~\cite{sundara_raman_censored_2020,niaki2020iclab}. 
Second, following the approach adopted by prior works, we shall cluster
the test records, such as, those predicted as probable censorship events, 
and manually examine each cluster to determine whether the prediction is correct
by inferring what the censorship mechanism is and what intention 
of censorship may be~\cite{jones2014automated,sundara_raman_censored_2020,niaki2020iclab}. 

\paragraph{E2ECD versus DenseNetCD}
We argue that the E2ECD model is superior to the DenseNetCD model; however, 
the E2ECD model is not without a limitation when compared to the DenseNetCD model.
The E2ECD model requires training two neural network models, a feature
representation learning model and a classification model. In contrast, the
DenseNetCD model only needs to train a single pre-trained DenseNet. We observe
that training the two networks of the E2ECD model requires inherently more
computation than training a single network of DenseNetCD. The E2ECD model took
far more iterations to train, more than 100 times the number of steps as our
DenseNetCD model. In addition, the Censor2Seq component of the E2ECD model is
far more complex and so requires more computation per training step. However,
we argue that this additional complexity is clearly worth the computation cost
given the striking difference between the abilities of the two models as 
discussed in Section~\ref{sec:eval:undermine}.

\paragraph{Replication Package}
Some unmitigated limitations and threats to validity discussed in the above necessitate 
a future exploration along those directions. Ultimately, the approach to examine 
these limitations and threats to
validity is via replication studies. To facilitate this, we compose a
replication package and make it publicly
available~\cite{shawn_p_duncan_censored_2022}.  This package can also be 
useful to serve as a research baseline, to extend our work,  and to help develop
fully automated censorship detection systems.

%% file: summ.tex
\section{Conclusion and Discussion}
\label{sec:summary}

We examine a design of a fully-automated network-based Internet censorship
detection system. There are primarily three challenges we must address: to
process an overwhelmingly large quantity of structured network measurement data for
censorship detection, to extract high quality features from this data, 
and to address the challenge of the lack of training
data for the proposed supervised learning approach.  Our contributions are
therefore as follows. 

First, we engineer an iterative data processing Python library that allows us
stream the network measurement data directly from its cloud storage. Second,
recognizing the importance of the structure and the order of elements in the
network measurement data, we design an unsupervised sequence-to-sequence deep
learning model to infer latent feature represents in embedding space from the
data. Third, we prepare a training dataset using known fingerprints of
network-based Internet censorship. Fourth, due to the lack of machine learning
approaches of network-based Internet censorship detection in the literature, we
build two complete censorship detection models, one based on feature
representation learning, and the other based on image classification, and
compare the predictive performance of the two models. As a result, the work
resulted in an ``end-to-end'' network-based Internet censorship detection
pipeline accompanied by a replication package publicly available for adoption
and extension.

\paragraph{Discussion} A worthy lesson is that there is a cyclic dependency or chicken-egg problem
when it comes to the design of the ``end-to-end'' censorship detection system.
Such a system shall be based on a supervised learning algorithm that requires a
large set of training network measurement test records labeled as ``censored''
or ``uncensored'' to learn to detect censorship; however, we need to be able
to recognize what constitutes censorship by examining these data records. As
a result, the cyclic dependency problem emerges. 

In this work, we leverage a set of censorship fingerprints in the form of
regular expressions and the censorship scanning scripts available by Censored
Planet.  The scripts use regular expressions to match received block page
responses, which requires discovering and analyzing such responses, then
designing a regular expression based on that analysis. This limits censorship
detection to patterns already discovered. However, our work indicates that the
labeled data set generated in this process is more than appropriate to serve as
the required training dataset. Indeed, our research affirms that a machine
learning model can learn to detect previously unseen block pages.  Both of our
two experimental models (E2ECD and DenseNetCD) detected block page responses
that do not match any of the regular expressions provided in the scanning
scripts from Censored Planet.  Further, our E2ECD model with
sequence-to-sequence latent feature representation learning identified as
probable censorship several types of responses that do not correlate to simple
textual patterns and therefore would be very difficult if not impossible to
detect using regular expression based scripts. Additionally, There is
a striking difference between the number of responses that the two deep
learning methods (E2ECD and DenseNetCD) identify as probable censorship. We
conclude that a sequence-to-sequence autoencoder enhanced with an attention 
mechanism as proposed is far
more effective at discovering censorship than the simple image feature
representation method used by DenseNetCD.

An added benefit of our research is that our results show that useful latent feature
representation of censorship measurement in embedding space can be identified
by deep neural networks. We expect that the embeddings should be sufficient to
identify new and known network-based censorship events. This is evident, as
shown in our evaluation, the two experimental networks displayed an accuracy
better than 98\% when processing responses that are already categorized by the
scanning scripts.  More significantly, both of our networks
discovered censorship missed by these scripts, showing that deep neural
networks can detect features not readily apparent.  Although machine learning
is no panacea, we are convinced that these deep learning networks have the
potential to be a powerful tool to understand the constantly changing landscape
of Internet censorship.

In addition, we only evaluate our model using the Censored Planet data set.  As
discussed in Section~\ref{sec:related}, there are now several complementary
prominent large scale network measurement platforms for Internet censorship.
Our future work includes the extension of the current model to handle and correlate
both ordinary network traffic measurement and network side-channel measurement
data from these complementary platforms. Additionally, there is significant and continuous
development in modeling techniques. We expect that our models can be improved using these
developments.

%% file: ack.tex
\section*{Acknowledgement}
This research was supported, in part, by a grant of computer time from the City
University of New York High Performance Computing Center.
This research was also supported, in part, by a grant of software as a service
from Comet ML (\url{https://comet.ml}).